# Word sense disambiguation criteria: a systematic study

**Laurent AUDIBERT**
DELIC, Université de Provence
29 Av. Robert SCHUMAN – 13621
Aix-en-Provence Cedex 1, FRANCE
`laurent.audibert@up.univ-aix.fr`

**Abstract**

This article describes the results of a systematic in-depth study of the criteria used for word sense disambiguation. Our study is based on 60 target words: 20 nouns, 20 adjectives and 20 verbs. Our results are not always in line with some practices in the field. For example, we show that omitting non-content words decreases performance and that bigrams yield better results than unigrams.

## 1 Introduction

The task of word sense disambiguation (WSD) is to identify the correct sense of a word in context. WSD is usually performed by matching information from the context in which the word occurs with disambiguation knowledge source. Our approach uses supervised machine-learning techniques to automatically acquire such disambiguation knowledge from sense-tagged corpora. At present, this type of approach is widely used and seems to yield the best results (Kilgarriff, Rosenzweig, 2000; Ng, 1997b).

Information conveyed by context words (morphological form) is enriched with further annotations: part-of-speech tag, lemma, etc. Each individual piece of information is called a feature. A good feature should capture an important source of knowledge that is critical in determining the sense of the word to be disambiguated. The choice of the appropriate set of features is an important issue for WSD (Bruce, Wiebe, Perdersen, 1996; Ng, Zelle, 1997; Pedersen, 2001). Thus, this paper describes the results of a systematic and in-depth study of homogenous criteria (*i.e.* set of features) that can be used for WSD.

## 2 Methodology

### 2.1 Corpus

The corpus we worked on is composed of different types of texts and comprises 6 468 522 words. It was put together within the framework of the SyntSem project that aims at producing a French corpus which is morphologically and syntactically tagged, lemmatised and that comprises a light syntactical tagging as well as a lexical tagging of 60 target words selected for their strongly polysemous nature (Véronis, 1998)[1]. These 60 target words are evenly divided into 20 nouns, 20 adjectives and 20 verbs, having a total of 53 796 occurrences in the corpus.

The inadequacy of standard dictionaries (Véronis, 2001) and computational lexicons (Palmer, 1998) for natural language processing is presently one of the major difficulties encountered in word sense disambiguation. For instance, by using these dictionaries, the inter-annotator agreement may sometimes reach only 57% (Ng, Lee, 1996) or may simply be equivalent to random sense allocation (Véronis, 1998). To overcome this weakness, a dictionary more specific to natural language processing is being developed in our team (Reymond, 2002). It has been used to tag the occurrences of the 60 target words of the SyntSem corpus.

Table 8 in the appendix gives quantitative information for each target word. The number of senses per word may be very large for it includes idioms and phrasal verbs such as: « mettre sur pied », « mettre à pied », « pied de nez », etc.

A general agreement seems to emerge according to which morpho-syntatic disambiguation and sense disambiguation can be disentangled (Kilgarriff, 1997; Ng, Zelle, 1997). We have entrusted the part-of-speech tagging of our corpus to the Cordial software (developed by Synapse Développement company) as it offers lemmatisation and part-of-speech tagging of a satisfactory accuracy (Valli, Véronis, 1999).

| mform | lemma | ems | cgems | sense |
|---|---|---|---|---|
| mettre | mettre | VINF | VINF | 1.12.7 |
| fin | fin | NCFS | NCOM | |
| à | à | PREP | PREP | |
| la | le | DETDFS | DET | |
| pratique | pratique | NCFS | NCOM | |
| des | de | DETDPIG | DET | |
| détentions | détention | NCFP | NCOM | 1 |

Table 1: SyntSem tagged corpus extract.

---

[1] These words are those used in the French part of the Senseval-1 evaluation exercice (Segond, 2000) but the corpus and dictionary are different in the present study.

Table 1 displays an extract of the SyntSem corpus. It shows all the tags of each word. We use the information provided by these tags in our lexical disambiguation criteria.

**2.2 Criteria**

The aim of our study is to evaluate a large variety of homogenous criteria (*i.e.* set of features). The name of each criterion specifies its nature and takes the following form *[par1|par2|par3|par4]*. Parameter *par1* indicates whether the criterion takes into account unigrams (*par1=1gr*), bigrams (*par1=2gr*) or trigrams (*par1=3gr*), knowing that an n-gram represents the juxtaposition of *n* words. Parameter *par2* indicates which word tag is considered: morphological form (*par2=mform*), lemma (*par2=lemma*), part-of-speech (*par2=ems*) or coarse-grained part-of-speech (*par2=cgems*). Parameter *par3* indicates if we take into account word positions (*par3=position*), if we only distinguish left from right context (*par3=leftright*), or if we simply consider unordered set of surrounding words (*par3=unordered*). Lastly, parameter *par4* shows whether the criterion takes into account all the words (*par4=all*) or content words only (*par3=content*). We call these criteria "homogeneous criteria" as the four parameters together determine the nature of all pieces of contextual evidence selected by the criterion.

For contexts within a range of ±1 to ± 8 words, the combination of all parameters generates 576 (3×4×3×2×8) distinct criteria. We have systematically evaluated each one of these criteria as well as other criteria in order to answer specific questions and to validate or invalidate certain hypothesis.

Within the framework of this study, we have developed an application used to model these criteria and to further apply them to the corpus in order to generate feature vectors used by our classifiers (Audibert, 2001).

**2.3 Classifiers**

We have selected two complementary classifiers. We have chosen the Naïve-Bayes classifier (NB) for its simplicity and widespread use, as well as for its well-known state-of-the-art accuracy on supervised WSD (Domingos, Pazzani, 1997; Mooney, 1996; Ng, 1997a). The NB classifier assumes the features are independent given the sense. During classification, it chooses the sense with the highest posterior probability. We have also selected a decision list classifier (DL) which is similar to the classifier used by (Yarowsky, 1994) for words having two senses, and extended for more senses by (Golding, 1995). DL classifier is further developed in (Audibert, 2003). In DL, features are sorted in order of decreasing strength, where strength reflects feature reliability for decision-making. The DL classifier distinguishes itself clearly from the NB classifier as it does not combine the features, but bases its classifications solely on the single most reliable feature identified in the target context selected by the criteria. We will make a use of this decision-making transparency several times in this article. Some other advantages of DL classifier are its significant simplicity and its ease of implementation.

Both of the classifiers we used require probability estimates. Given the data-sparseness, we have to deal with zero or low frequency counts. For this reason, we have decided to use m-estimation (Cussens, 1993) rather than classical estimations of probabilities or Laplace ("add one") smoothing.

When a classifier is not able to disambiguate a target word, which is very rare, it selects the most frequent sense from the training data. Thus, all occurrences are tagged. As in this case precision equals the recall, the present article speaks in terms of precision only.

To evaluate a criterion in the corpus, we use a k-fold cross-validation method (in accordance with the common use, in our experiment, k=10). Despite the fact that this method takes much computing time, it enables the evaluation of the criterion in the whole corpus.

Throughout the tests, the two classifiers have generally obtained comparable accuracy, although the NB classifier has almost systematically outperformed the DL classifier.

**3 Results**

**3.1 Best criteria precision**

Table 2 displays for each of the target words studied the optimal context size and the disambiguation precision obtained by the best unigram, bigram and trigram-based criteria.

This table points out that best criteria take into account all words in the context. Section 3.2 will concentrate on feature reliability according to their part-of-speech. Then, section 3.2 will deal with the impact of different feature selections based on features part-of-speech.

|  | Nouns | | Adjectives | | Verbs | |
|---|---|---|---|---|---|---|
| **Criterion** | **P%** | **S** | **P%** | **S** | **P%** | **S** |
| [1gr|lemma|ordered|all] | 81.9 | ±2 | 76.8 | ±1 | 71.8 | ±3 |
| [2gr|lemma|leftright|all] | 83.6 | ±4 | 77.9 | ±3 | 74.0 | ±4 |
| [3gr|lemma|leftright|all] | 82.3 | ±5 | 72.7 | ±3 | 71.2 | ±5 |

Table 2: Optimal context size (S) and criteria precision (P%) using NB classifier.

According to the Table 2, the optimal context size comprises ±1 to ±5 words. Further developments of the context optimality will be made in section 3.4.

Surprisingly, Table 2 outlines the fact that the criterion that obtains the best precision is based on bigrams and not on unigrams. This subject is dealt with in section 3.5.

### 3.2 The most reliable parts-of-speech

In this section, we aim at learning the part-of-speech and the space distribution of all pieces of contextual evidence used for disambiguation.

To this end, we use the DL classifier because it bases its classifications solely on the most reliable piece of evidence identified by the criteria. Thus, DL classifier enables us to learn which is the part-of-speech and the space distribution of this indicator by using the criterion *[1gr/mform/ordered/all]*. Table 3 and graphics presented in Figure 1 synthesize this study results.

Table 3 enables to bring out the following results (we quote between brackets and in order the main results obtained for the nouns, the adjectives and the verbs):

- common nouns (NCOM) obtain one of the best precisions (93.0%; 93.7%; 87.8%) and represent one of the most used indicators (12.7%; 25.3%; 26%) for the three term categories;
- adjectives (ADJ) represent good indicators for nouns (p=94.9%) and adjectives (p=80.3%), but they are especially useful for nouns since they are used in 13.7% of the cases against 2.2% only for the adjectives;
- adverbs (ADV) are mainly useful for adjectives; their precision reaches 79.2% and they are used in 9.6% of the cases;
- verbs in the infinitive (VINF) are very reliable indicators for the three parts-of-speech (90.2%; 80.6%; 87.2%), but they are rarely used as they are not very often encountered in the context (0.9%; 0.7%; 2.9%);
- conjugated verbs (VCON) obtain poor precision (67.9%; 53.9%; 54.4%).

Figure 1 graphics show the space distribution of the main parts-of-speech of the indicators used to disambiguate each one of the three term categories. The dissymmetric shape of verbs, and more precisely, the strong prevalence of indicators located in position +1, +2, +3, makes us believe that disambiguating verbs is done more accordingly to their object than to their subject since as a rule the main form encountered is *subject–verb–object*.

|  | Nouns | | Adjectives | | Verbs | |
|---|---|---|---|---|---|---|
|  | P% | U% | P% | U% | P% | U% |
| **NCOM** | 93.0 | 12.7 | 93.7 | 25.3 | 87.7 | 26.0 |
| **DET** | 73.8 | 30.2 | 69.8 | 21.3 | 48.1 | 12.7 |
| **PREP** | 78.3 | 24.2 | 61.4 | 15.2 | 62.9 | 17.4 |
| **ADJ** | 94.9 | 13.7 | 80.3 | 2.2 | 65.5 | 3.2 |
| **ADV** | 57.0 | 1.1 | 79.2 | 9.6 | 60.3 | 5.7 |
| **PROPE** | 63.6 | 0.6 | 67.0 | 2.6 | 65.9 | 11.4 |
| **PCTFORTE** | 72.1 | 3.00 | 69.0 | 4.2 | 80.9 | 2.9 |
| **VCON** | 67.9 | 1.5 | 53.9 | 2.4 | 54.4 | 3.0 |
| **SUB** | 78.6 | 0.6 | 58.1 | 1.8 | 79.9 | 2.7 |
| **VINF** | 90.2 | 0.9 | 80.7 | 0.7 | 87.2 | 2.9 |
| **NPRO** | 86.8 | 0.3 | 92.0 | 0.6 | 81.8 | 1.2 |
| **VPAR** | 89.7 | 0.3 | 50.0 | 0.2 | 81.0 | 0.9 |
| **PRODE** | 100 | 0.0 | 35.0 | 0.2 | 68.6 | 0.8 |

Table 3: Precision (P%) and usage proportion (U%) by coarse-grained part-of-speech of most reliable contextual evidences using DL classifier.

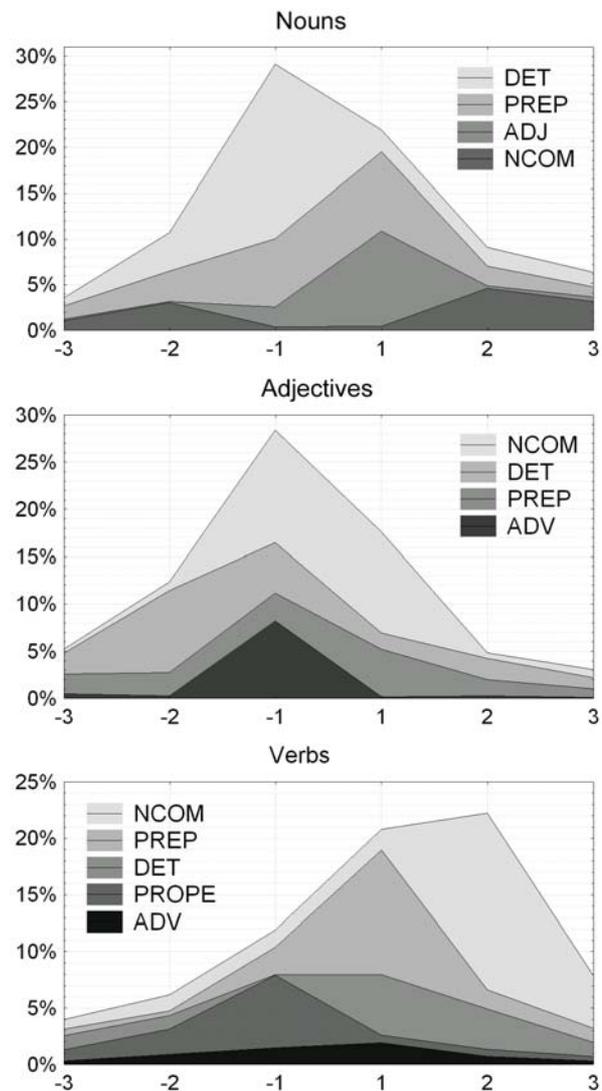

Figure 1: space distribution by part-of-speech of most reliable pieces of contextual evidence used for disambiguation with DL classifier.

|         | Most reliable contextual evidences |     |     |     |      |
|---------|------|-----|-----|------|-------|
|         | NCOM | ADJ | ADV | DET  | PREP  |
| Nouns   | -2, +2 | +1 |   | -1   | -1, +1 |
| Adject. | -1, +1 |    | -1 | -1, -2 | +1  |
| Verbs   | +2, +3 |    |    | +1, +2 | +1  |

Table 4: space distribution of most reliable pieces of contextual evidence used for disambiguation with DL classifier.

Table 4 summarizes these graphs. Our results and those of (Yarowsky, 1993) agree in many respects, although his study applies only to pseudo-words having only two "senses":

- "Adjectives derive almost all of their disambiguating information from the nouns they modify";
- "Nouns are best disambiguated by directly adjacent adjectives or nouns";
- "Verbs derive more disambiguating information from their objects than from their subjects".

### 3.3 The importance of stop-words

|          | Nouns | Adjectives | Verbs |
|----------|-------|------------|-------|
| Unigrams | 0.3%  | 2.5%       | 6.9%  |
| Bigrams  | 2.7%  | 3.4%       | 13.5% |
| Trigrams | 12.4% | 15.9%      | 20.2% |

Table 5: precision decrease when omitting non-content words using DL classifier.

Many studies do not consider all the words of the context (El-Bèze, Loupy, Marteau, 1998; Mooney, 1996; Ng, Lee, 2002). The assumption according to which content words represent the most reliable indicators underlies the choice to use only content words based criteria. This seems to be obvious, but it is not confirmed in Table 2. Table 5 shows the average decrease of the precision of the content words based criteria (*[par1|par2|par3|content]*) compared to the same criteria based on all words (*[par1|par2|par3|all]*). This table shows that the decrease is low when the criteria are based on unigrams and are used to disambiguate nouns, but it can become very high in the other cases, and in particular for verbs disambiguation.

Table 3 informs us about the disambiguation precision according to the coarse-grained part-of-speech tag. This table shows that using content words only is probably not the most appropriate feature selection (for example inflected verbs are not relevant). We have therefore chosen to try out a more subtle selection (we refer to it by *par4=selected*) by restraining to the most reliable parts-of-speech according to Table 3:

- For nouns, we use indicators having the following coarse-grained part-of-speech tagging:

|                              | Nouns | Adj. | Verbs |
|------------------------------|-------|------|-------|
| [1gr|mform|ordered|all]      | 81.5  | 75.7 | 71.0  |
| [1gr|mform|ordered|content]  | 78.9  | 71.6 | 59.5  |
| [1gr|mform|ordered|selected] | 79.2  | 71.5 | 66.3  |

Table 6: precision with and without feature selections using NB classifier.

NCOM, PREP, ADJ, SUB, VINF, NPRO, VPAR or PRODE;
- For adjectives, we use indicators having the following coarse-grained part-of-speech tagging: NCOM, DET, ADJ, ADV, VINF or NPRO;
- For verbs, we use indicators having the following coarse-grained part-of-speech tagging: NCOM, ADJ, PROPE, PCTFORTE, SUB, VINF, NPRO, VPAR or PRODE.

Table 6 gives a comparison of the precision reached by the following 3 criteria:
- *[1gr|mform|ordered|all]*,
- *[1gr|mform|ordered|content]*,
- *[1gr|mform|ordered|selected]*.

We observe that this subtler selection lowers the disambiguation precision too. We assume then that all words, whatever their part-of-speech, contribute to the disambiguation.

### 3.4 Optimal context

#### 3.4.1 Size and symmetry

We tested up to ±10 000 word contexts. However, the best precision is always obtained for short contexts ranging from ±1 to ±5 words. These results are similar to those obtained by many other researches (El-Bèze, et al., 1998; Yarowsky, 1993; 2000).

Optimal context size is criteria, target part-of-speech and n-gram size dependent. In particular, it increases with the n-gram size.

Table 7 shows, for all the criteria we examined, the average size of the optimal context by the n-gram size and by the part-of-speech.

The main indicators used to disambiguate nouns and adjectives surround roughly symmetrically the word we want to disambiguate. On the contrary, indicators for verbs tend to be mainly situated after the verb. Therefore, a non-symmetrical context shifted forward by a word proves to be more appropriate. Our experiments show that the use of this shifted context improves the precision of the verbs disambiguation by 0.75% in average.

|          | Nouns | Adjectives | Verbs |
|----------|-------|------------|-------|
| Unigrams | 1.5   | 1.1        | 1.8   |
| Bigrams  | 2.4   | 2          | 2.8   |
| Trigrams | 3.1   | 3.4        | 3.8   |

Table 7: optimal context size using both classifiers.

### 3.4.2 Do n-grams have to contain the target word?

The lemma being unique for a given word, if only lemmas are considered, an n-gram which is adjacent to the target word contains precisely the same information as the same n-gram to which the target word is added in order to compose a (n+1)-gram. The n-gram that is situated next to the word to disambiguate can thus be assimilated to the (n+1)-gram which contains it. Consequently, the question becomes: do n-grams have to contain the word to disambiguate or at least to be adjacent to it? Several studies set themselves this constraint probably because n-grams are used to capture fixed constructions containing the word to disambiguate.

Table 2 shows that the optimal context size for best bigram or trigram-based criteria does not fit this constraint. The relevant n-grams do not necessarily contain the target word and are not necessarily adjacent to it. For example, for nouns and verbs, the ±4 words context is the optimal context size of the bigram-based criteria which obtains the best disambiguation precision. This criterion produces some bigrams separated from the target word by one or two words. However, this single observation cannot enable us to abandon the constraint in terms of containing or being adjacent to the target word. Indeed, the bigram increasing distance may help locating an information which could be captured by the joint use of one or several larger n-grams. We have thus evaluated a combination of criteria in which all n-grams contain the target word in a context up to ±4 words:

- [2gr|lemma|leftright|all] with context size of ±1 words;
- [3gr|lemma|leftright|all] with context size of ±2 words;
- [4gr|lemma|leftright|all] with context size of ±3 words;
- [5gr|lemma|leftright|all] with context size of ±4 words.

This combination leads to a disambiguating precision of 74.3%, which is lower than the one obtained using the criteria *[2gr|lemma|leftright|all]* alone with a ±4 words context. This experiment confirms that constraining the context to contain the word to disambiguate, or at least to be adjacent to it, decreases disambiguation accuracy. Consequently, nothing justifies this constraint on criteria.

### 3.5 Surprising bigrams accuracy

Contrary to all expectations, Table 2 shows that the best unigram-based criterion (*[1gr|lemma|ordered|all]*) is definitely less accurate than the best bigram-based criterion (*[2gr|lemma|leftright|all]*). However, in practice, bigrams and trigrams are seldom used alone. When used, they are taken in conjunction with unigrams which are supposed to convey the most reliable piece of information.

Why does the criterion *[2gr|lemma|leftright|all]* work so well? First, since this criterion is a juxtaposition of lemmas, among the features generated by this criterion, the left and the right unigrams are to be found, even if these unigrams are disguised as bigrams (cf. section 3.4.2). As these pieces of contextual evidence are certainly the most important ones (cf. section 3.4), it makes sense that this bigram-based criterion obtains good results.

Second, in a context of a higher size, the juxtaposition of two words seems more relevant than one isolated word. For example, to disambiguate the word *utile*, the bigram *pour_le* is relevant, whereas the single unigrams *pour* and *le* are not of much help.

Lastly, sometimes, the presence of a unigram noncontiguous to the target word can be sufficient to solve the ambiguity. But using bigram-based criteria does not necessarily lose the piece of information conveyed by unigram-based criteria. For example, a determiner, a preposition or an apostrophe often precedes a common noun. The lemmatisation variability of this determiner, this preposition or this apostrophe is low for a given common noun located at a given distance from a given target word. Therefore, the piece of information brought out by the juxtaposition of the noun and the preceding word is often very similar to the piece of information provided by the noun alone.

## 4 Conclusion

We have described here the results of a systematic and in-depth research on WSD criteria. This may be the first research of this extent carried out within a unified framework. This study enabled us to confirm certain results stated in the field literature such as:

- importance of short contexts;
- importance of adjacent noun or adverb for adjective disambiguation;
- importance of adjacent adjective, or noun in a very short context for noun disambiguation;
- importance of the noun in the area after the verb and use of dissymmetrical contexts for verb disambiguation.

We have also obtained more original results, not always in line with some practices in the field such as:

- importance of stop-words whose withdrawal decreases the performance almost systematically;
- better results obtained by bigrams taken alone than unigrams alone;
- unnecessary constraint of including or be adjacent to the target word.

Disambiguation accuracy could probably be improved by the study of other sources of information useful in disambiguation, such as:

- criteria based on binary syntactic relations (noun-noun, noun-verb, adjective-noun, etc.) to capture information which can be absent from short contexts;
- the use of thesauri or other sources of information to carry out generalizations on context words to overcome data sparseness problem;
- topical text information;
- selectional restrictions.

This preliminary study focuses on homogenous criteria (for example: lemmas located from –2 to +2 position). To improve the disambiguation accuracy, we have to look for heterogeneous criteria by gathering the most relevant pieces of contextual evidence not necessarily of the same type (for example: lemma in position –2, part-of-speech in position –1, morphological form of target word and lemma in position +2). This feature selection leads to a combinatorial explosion that can be solved by genetic algorithms (Daelemans, Hoste, Meulder, Naudts, 2003).

**References**

Audibert L. (2001), LoX: Outil Polyvalent pour l'Exploration de Corpus Annotés, *5ème Rencontre des étudiants Chercheurs en Informatique pour le Traitement Automatique des Langues (RECITAL-2001)*, 411-419.

Audibert L. (2003), Etude des Critères de Désambiguïsation Sémantique Automatique: Résultats sur les Cooccurrences, *10ème conférence sur le Traitement Automatique des Langues Naturelles (TALN-2003)*, 35-44.

Bruce R., Wiebe J., Perdersen T. (1996), The Measure of a Model, *1st Conference on Empirical Methods in Natural Language Processing (EMNLP-1996)*, 101-112.

Cussens J. (1993), Bayes and Pseudo-Bayes Estimates of Conditional Probability and their Reliability, *6th European Conference on Machine Learning (ECML-1993)*, 136-152.

Daelemans W., Hoste V., Meulder F. D., Naudts B. (2003), Combined Optimization of Feature Selection and Algorithm Parameter Interaction in Machine Learning of Language, *14th European Conference on Machine Learning (ECML-2003)*, 84-95.

Domingos P., Pazzani M. (1997), Beyond Independence: Conditions for the Optimality of the Simple Bayesian Classifier, *Machine Learning*, 29: 103-130.

El-Bèze M., Loupy C. d., Marteau P.-F. (1998), WSD Based on Three Short Context Methods, *SENSEVAL Workshop*, in press.

Golding A. R. (1995), A Bayesion Hybrid Method for Context-Sensitive Spelling Correction, *3th Workshop on Very Large Corpora*, 39-53.

Kilgarriff A. (1997), Evaluating Word Sense Disambiguation Programs: Progress Report, *Speech and Language Technology (SALT-1997) Workshop on Evaluation in Speech and Language Technology*, 114-120.

Kilgarriff A., Rosenzweig J. (2000), English Senseval: Report and Results, *2nd International Conference on Language Resources and Evaluation (LREC-2000)*, 3: 1239-1244.

Mooney R. J. (1996), Comparative Experiments on Disambiguating Word Senses: an Illustration of the Role of Bias in Machine Learning, *1st Conference on Empirical Methods in Natural Language Processing (EMNLP-1996)*, 82-91.

Ng H. T. (1997a), Exemplar-Based Word Sense Disambiguation: Some Recent Improvements, *2nd Conference on Empirical Methods in Natural Language Processing (EMNLP-1997)*, 208-213.

Ng H. T. (1997b), Getting Serious About Word Sense Disambiguation, *Association for Computational Linguistics Special Interest Group on the Lexicon (ACL-SIGLEX-1997): Workshop "Tagging Text with Lexical Semantics: Why, What, and How ?"* 1-7.

Ng H. T., Lee Y. K. (1996), Integrating Multiple Knowledge Sources to Disambiguate Word Sense: An Exemplar-Based Approach, *34th Annual Meeting of the Society for Computational Linguistics*, 40-47.

Ng H. T., Lee Y. K. (2002), An Empirical Evaluation of Knowledge Sources and Learning Algorithms for Word Sense Disambiguation, *7th Conference on Empirical Methods in Natural Language Processing (EMNLP-2002)*, 41-48.

Ng H. T., Zelle J. (1997), Corpus-Based Approaches to Semantic Interpretation in Natural Language Processing, *Artificial Intelligence Magazine - Special Issue on Natural Language Processing*, 18: 45-64.

Palmer M. (1998), Are WordNet Sense Distinctions Appropriate for Computational

Lexicons, *Association for Computational Linguistics Special Interest Group on the Lexicon (ACL-SIGLEX-1998): Senseval*, in press.

Pedersen T. (2001), Machine Learning with Lexical Features: the Duluth Approach to Senseval-2, *2nd International Workshop on Evaluating Word Sense Disambiguation Systems (Senseval-2)*, 139-142.

Reymond D. (2002), Méthodologie pour la Création d'un Dictionnaire Distributionnel dans une Perspective d'Étiquetage Lexical Semi-Automatique, *6ème Rencontre des étudiants Chercheurs en Informatique pour le Traitement Automatique des Langues (RECITAL-2002)*, 405-414.

Segond F. (2000), Framework and Results for French, *Computers and the Humanities*, 34: 49-60.

Valli A., Véronis J. (1999), Étiquetage Grammatical de Corpus Oraux: Problèmes et Perpectives, *Revue Française de Linguistique Appliquée*, 4: 113-133.

Véronis J. (1998), A Study of Polysemy Judgements and Inter-Annotator Agreement, *Programme and Advanced Papers of the Senseval-1 Workshop*, 2-4.

Véronis J. (2001), Sense Tagging: Does It Make Sense ?, *Corpus Linguistics Conference*, http://www.up.univ-mrs.fr/~veronis/pdf/2001-lancaster-sense.pdf.

Yarowsky D. (1993), One Sense Per Collocation, *ARPA Workshop on Human Language Technology*, 266-271.

Yarowsky D. (1994), A Comparision of Corpus-Based Techniques for Restoring Accents in Spanish and French Text, *2nd Annual Workshop on Very Large Text Corpora*, 19-32.

Yarowsky D. (2000), Hierarchical Decision List for Word Sense Disambiguation, *Computers and the Humanities*, 34: 179-186.

## Appendix

| Nouns | F | S | H | MFS |
|---|---|---|---|---|
| barrage | 92 | 5 | 1.2 | 76.1% |
| chef | 1133 | 11 | 1.5 | 76.0% |
| communication | 1703 | 13 | 2.4 | 40.6% |
| compagnie | 412 | 12 | 1.6 | 71.4% |
| concentration | 246 | 6 | 2 | 45.1% |
| constitution | 422 | 6 | 1.6 | 50.0% |
| degré | 507 | 18 | 2.5 | 58.6% |
| détention | 112 | 2 | 0.9 | 72.3% |
| économie | 930 | 10 | 2.2 | 49.1% |
| formation | 1528 | 9 | 1.7 | 63.8% |
| lancement | 138 | 5 | 1 | 79.7% |
| observation | 572 | 3 | 0.7 | 86.0% |
| organe | 366 | 6 | 2.2 | 38.3% |
| passage | 600 | 19 | 2.7 | 37.0% |
| pied | 960 | 62 | 3.5 | 37.6% |
| restauration | 104 | 5 | 1.8 | 43.3% |
| solution | 880 | 4 | 0.4 | 93.3% |
| station | 266 | 8 | 2.6 | 32.0% |
| suspension | 110 | 5 | 1.5 | 61.8% |
| vol | 278 | 10 | 2.2 | 40.3% |
| **Average** | **568** | **14.2** | **1.9** | **57.3%** |
| **Adjectives** | **F** | **S** | **H** | **MFS** |
| biologique | 475 | 4 | 0.5 | 89.9% |
| clair | 557 | 20 | 3.1 | 29.3% |
| correct | 116 | 5 | 1.8 | 53.4% |
| courant | 170 | 4 | 0.6 | 90.0% |
| exceptionnel | 226 | 3 | 1.4 | 53.1% |
| frais | 184 | 18 | 3.1 | 36.4% |
| haut | 1017 | 29 | 3.5 | 25.0% |
| historique | 620 | 3 | 0.7 | 87.4% |
| plein | 844 | 35 | 4 | 17.1% |
| populaire | 457 | 5 | 2 | 47.9% |
| régulier | 181 | 11 | 2.5 | 32.6% |
| sain | 129 | 10 | 2.4 | 40.3% |
| secondaire | 195 | 5 | 1.7 | 53.8% |
| sensible | 425 | 11 | 2.6 | 29.9% |
| simple | 1051 | 14 | 2.1 | 41.3% |
| strict | 220 | 9 | 2.2 | 45.5% |
| sûr | 645 | 14 | 2.6 | 45.9% |
| traditionnel | 447 | 2 | 0.5 | 89.5% |
| utile | 359 | 9 | 2.4 | 42.9% |
| vaste | 368 | 6 | 2.1 | 42.4% |
| **Average** | **434** | **14.1** | **2.3** | **46.4%** |
| **Verbs** | **F** | **S** | **H** | **MFS** |
| arrêter | 916 | 15 | 3 | 23.9% |
| comprendre | 2145 | 13 | 2.8 | 32.6% |
| conclure | 727 | 16 | 2.4 | 45.5% |
| conduire | 1093 | 15 | 2.3 | 38.2% |
| connaître | 1635 | 16 | 2.2 | 40.1% |
| couvrir | 543 | 22 | 3.3 | 33.3% |
| entrer | 1258 | 39 | 3.7 | 26.6% |
| exercer | 698 | 8 | 1.5 | 59.5% |
| importer | 576 | 8 | 2.6 | 27.6% |
| mettre | 5246 | 140 | 3.7 | 42.2% |
| ouvrir | 919 | 41 | 3.8 | 26.0% |
| parvenir | 654 | 8 | 2.3 | 36.7% |
| passer | 2556 | 84 | 4.5 | 15.8% |
| porter | 2347 | 59 | 4 | 29.4% |
| poursuivre | 978 | 16 | 2.7 | 36.2% |
| présenter | 2142 | 18 | 2.6 | 40.1% |
| rendre | 1990 | 27 | 2.9 | 46.4% |
| répondre | 2529 | 9 | 1 | 78.3% |
| tirer | 1002 | 47 | 3.9 | 28.9% |
| venir | 3797 | 33 | 3.2 | 24.9% |
| **Average** | **1688** | **47.4** | **3.1** | **37.2%** |

Table 8: target word frequency (F), average number of senses (S), sense repartition entropy (H) and base-line accuracy (Most Frequent Sense: MFS).